\documentclass[nonacm,acmlarge,screen]{acmart}

\AtBeginDocument{%
  \providecommand\BibTeX{{%
    \normalfont B\kern-0.5em{\scshape i\kern-0.25em b}\kern-0.8em\TeX}}}

\usepackage{graphicx}
\usepackage{array}
\usepackage{cleveref}

\usepackage{caption}
\usepackage{subcaption}
\usepackage{rotating}
\usepackage{lscape}

\begin{document}
\title{A Collaborative, Human-Centred Taxonomy of AI, Algorithmic, and Automation Harms}

\authorsaddresses{}

\author{Gavin Abercrombie}\authornotemark[1]
\affiliation{
  \institution{Heriot-Watt University}
  \city{Edinburgh}
  \country{Scotland}
}

\author{Djalel Benbouzid}\authornotemark[1]
\affiliation{
    \institution{Machine Learning Research Lab, Volkswagen Group}
    \city{Munich}
    \country{Germany}
    }
\email{djalel@argmax.ai}

\author{Paolo Giudici}\authornotemark[1]
\affiliation{
  \institution{University of Pavia \& European University Institute}
  \city{Pavia}
  \country{Italy}
}

\author{Delaram Golpayegani}\authornotemark[1]
\affiliation{
  \institution{ADAPT Centre, Trinity College Dublin}
  \city{Dublin}
  \country{Ireland}
}

\author{Julio Hernandez}\authornotemark[1]
\affiliation{
  \institution{ADAPT Centre, Trinity College Dublin}
  \city{Dublin}
  \country{Ireland}
}

\author{Pierre Noro}\authornotemark[1]
\affiliation{
  \institution{Tech \& Global Affairs Hub, SciencesPo Paris}
  \city{Paris}
  \country{France}
}

\author{Harshvardhan Pandit}\authornotemark[1]
\affiliation{
  \institution{ADAPT Centre, Dublin City University}
  \city{Dublin}
  \country{Ireland}
}

\author{Eva Paraschou}\authornotemark[1]
\affiliation{
  \institution{Aristotle University of Thessaloniki}
  \city{Thessaloniki}
  \country{Greece}
}

\author{Charlie Pownall} %
\affiliation{
  \institution{AIAAIC}
  \city{Cambridge}
  \country{UK}
}

\author{Jyoti Prajapati}\authornote{Author ranking is alphabetical.}
\affiliation{
  \institution{TEC, Department of Telecommunications, Government of India}
  \city{New Delhi}
  \country{India}
}

\author{Mark A. Sayre}\authornotemark[1]
\affiliation{
  \institution{University of Maine School of Law}
  \city{Portland}
  \country{USA}
}

\author{Ushnish Sengupta}\authornotemark[1]
\affiliation{
  \institution{Algoma University}
  \city{Brampton}
  \country{Canada}
}

\author{Arthit Suriyawongkul}\authornotemark[1]
\affiliation{
  \institution{ADAPT Centre, Trinity College Dublin}
  \city{Dublin}
  \country{Ireland}
}

\author{Ruby Thelot}\authornotemark[1]
\affiliation{
  \institution{New York University}
  \city{New York}
  \country{USA}
}

\author{Sofia Vei}\authornotemark[1]
\affiliation{
  \institution{Aristotle University of Thessaloniki}
  \city{Thessaloniki}
  \country{Greece}
}

\author{Laura Waltersdorfer}\authornotemark[1]
\affiliation{
  \institution{TU \& WU Wien}
  \city{Vienna}
  \country{Austria}
}

\begin{abstract}
\label{sec:abstract}
This paper introduces a collaborative, human-centred taxonomy of AI, algorithmic and automation harms. We argue that existing taxonomies, while valuable, can be narrow, unclear, typically cater to practitioners and government, and often overlook the needs of the wider public.
Drawing on existing taxonomies and a large repository of documented incidents, we propose a taxonomy that is clear and understandable to a broad set of audiences, as well as being flexible, extensible, and interoperable.
Through iterative refinement with topic experts and crowdsourced annotation testing, we propose a taxonomy that can serve as a powerful tool for civil society organisations, educators, policymakers, product teams and the general public.
By fostering a greater understanding of the real-world harms of AI and related technologies, we aim to increase understanding, empower NGOs and individuals to identify and report violations, inform policy discussions, and encourage responsible technology development and deployment.
\end{abstract}

\maketitle

\section{Introduction}
\label{sec:introduction}

Much has been said about the benefits, opportunities and risks of artificial intelligence (AI), algorithms and automation, notably by industry, government and advisors. Conversely, researchers and others have spent significantly less time exploring and assessing the near-term harms of these technologies, even though these are real, can be highly damaging, and affect individuals, groups, communities, organisations, societies and environmental ecosystems~\citep{hanna-bender-2024-theoretical}.

On the surface, this seems surprising. The harms of these systems are wide-ranging, and significant. They have been known to result in the erosion and loss of fundamental liberties, such as through wrongful arrests enabled by flawed facial recognition technology\footnote{\url{https://apnews.com/article/technology-louisiana-baton-rouge-new-orleans-crime-50e1ea591aed6cf14d248096958dccc4}},
and infringe on the well-being and livelihoods of artists and other creators, whose works are used without acknowledgment or consent to train large language models\footnote{\url{https://www.mondaq.com/china/copyright/1444468/china-issues-landmark-ruling-on-copyright-infringement-involving-ai-generated-images}}.
Furthermore, they create and exacerbate social problems like teenage addiction to social media platforms fueled by attention-seeking algorithms\footnote{\url{https://www.wired.com/story/how-a-british-teens-death-changed-social-media}},
disrupt democratic political elections in the form of deepfakes\footnote{\url{https://news.sky.com/story/fake-ai-generated-joe-biden-robocall-tells-people-in-new-hampshire-not-to-vote-13054446}},
and place immense strain on local communities by depleting essential resources such as energy and water through the operation of data centres\footnote{\url{https://www.theatlantic.com/technology/archive/2024/03/ai-water-climate-microsoft/677602/}}.
These are just a few examples of the diverse and far-reaching harms that AI and AI-adjacent technologies are known to inflict.

Nonetheless, the identification and evaluation of negative impacts can be challenging. The workings of some technological systems, notably deep neural networks, remain opaque, partly due to inadequate explainability. More broadly, in the absence of legislation and reporting standards, system developers and deployers have little incentive to publicly acknowledge or report upstream or downstream issues, incidents or harms lest they increase their strategic, reputational, operational, financial or legal exposure.

Harms can also be hard to grasp and quantify; differentiating between the primary, secondary and tertiary impacts of particular technologies or applications, or their direct and indirect, and tangible and intangible impacts, is trickier than first meets the eye~\cite{cset}.
In addition, the harms of new technologies such as generative AI and emotion recognition are unclear and will evolve as new uses emerge, bad actors find novel ways to misuse them, and societal attitudes and behaviours evolve.
Further complicating this picture is the often unregulated marketing of poorly designed products~\cite{raji_fallacy_2022} that achieve mainstream adoption with low barriers to entry.

This information vacuum means most consumers and citizens, and many others, while aware of AI and related technologies, remain in the dark about the actual damage caused by their use and misuse--a lack of knowledge and understanding that has likely resulted in widespread and growing public scepticism about the use of these technologies in daily life~\cite{tyson-kikuchi-2023-growing}.\looseness=-1

Harms taxonomies can provide a solid foundation for informing policy, educating citizens, capturing and tracking violations, and managing product risks. A number of taxonomies of AI-related harms exist, but these tend to reflect the goals and biases of their creators, suffer from inconsistent nomenclature and definitions, mean different things to different people, and are incomplete and not maintained.

The UN AI Advisory Board has described developing ``a comprehensive list of AI risks for all time’ as a ‘fool’s errand''~\cite{UN2023}. It may be challenging, but the development of a clear, horizontal harms taxonomy is a worthwhile, indeed necessary, endeavour. It should help solidify the harms--and risks-- and thereby:

\begin{itemize}
    \item improve general public and other stakeholders' literacy;
    \item empower citizens and NGOs to report violations;
    \item strengthen the armoury of journalists, investigators and auditors~\citep{ojewale2024towards};
    \item put more power in the hands of civil society organisations;
    \item enable more effective risk management, and
    \item foster a greater sense of ethics and responsibility amongst developers, deployers, and others.
\end{itemize}

Prior works to develop taxonomies of AI harms~\cite{oecd_stocktaking_2023,nist_risk} are primarily designed to enable policymakers to inform legislative proposals, and develop risk assessment and mitigation strategies. These works may be beneficial and contribute to a better understanding and systematic analysis of harms, but they are often narrow, unclear in terms of categorisation, naming and definitions, and become quickly out-of-date.

Consequently, the great majority of consumers and citizens, as well as many students, journalists, policymakers, and others remain woefully uninformed about the real dangers and actual harms of AI and related technologies. Little surprise that the general public often fails to report harms when they occur, and is generally excluded from the risks and harms identification, classification and reporting process, and from the governance debate that should rightfully follow~\citep{cip-2023-participatory}.

With the focus of policy-makers and regulators shifting from risks to violations, and opinion against AI increasing in many countries as exposure increases\footnote{\url{https://www.politico.com/newsletters/digital-future-daily/2023/09/25/ai-vs-public-opinion-00118002}}, we believe it is an opportune time to introduce such a classification.

This paper introduces a collaborative and human-centred harms taxonomy aimed to serve as a powerful tool for civil society organisations, educators, policymakers, and product teams.
We draw on other taxonomies, topic experts, and the $1,500+$ incidents and issues captured in the AIAAIC Repository\footnote{\url{https://www.aiaaic.org/aiaaic-repository}} to set out a taxonomy of harms experienced across a wide variety of domains, countries, and cultures due to the inappropriate use and misuse of AI and adjacent technologies.

Clustered into nine top-level and sixty-nine sub-categories, our taxonomy has benefited from several iterations having been reviewed by experts from multiple disciplines and members of the general public from diverse backgrounds, genders, ages, and interests. It has also been tested multiple times using a custom-built annotation tool.

Our work is not intended to disrupt prior work, but to build on it in a way that makes the harms of AI and related systems clearer and more relevant and actionable to a broader set of audiences. As an ongoing process, we hope to add to the public discussion and technology ecosystem by building an understandable, practical, maintainable, open resource for humankind.

This paper is structured as follows:

\begin{itemize}
    \item Section 2 assesses the features and limitations of existing harms taxonomies
    \item Section 3 sets out the process by which the new taxonomy was developed and refined
    \item Section 4 describes the proposed new taxonomy
    \item Section 5 sets out ways the new taxonomy can be used; and
    \item Section 6 discusses the benefits and limitations of the new taxonomy and sets out potential next steps for its development.
\end{itemize}

\section{Related Work}
\label{sec:related_work}

\begin{table*}[ht!]
  \small
  \centering
  \renewcommand{\arraystretch}{1.2} %
  \setlength{\tabcolsep}{4pt} %
  \begin{tabular}{l p{4cm} p{2.5cm} p{2cm} p{3cm} }
  \hline
  \textbf{Taxonomy} & \textbf{Purpose / Focus} & \textbf{Target Audience} & \textbf{Publ. Type} & \textbf{Features}\\
  \\[-0.5ex]
  \hline
  \rule{0pt}{10pt}\textit{Generic scope} & & & & \\
  \midrule
  \citet{oecd_stocktaking_2023} & AI incident definition & Policy makers & White Paper & Harm dimensions (severity, types) \\
  \citet{cset} & AI incident analysis & AI community & White paper & Tangible vs. intangible harms \\
  \citet{nist_risk} & AI risk management & AI community & White paper & Technical design attributes \\
  \citet{alanturing} & Ethical AI design & Developers \& Project Staff & White Paper & Context of ethical platform \\
  \citet{microsoft} & Ethical AI design & Developers & Documentation & Development toolkit \\
  \midrule
  \rule{0pt}{10pt}\textit{Specific scope} & & & & \\
  \midrule
  \citet{shelby_sociotechnical_2023} & Socio-technical harms & AI community & Peer-reviewed & Survey on 172 papers \\
  \citet{raji_fallacy_2022} & AI system failures & AI community & Peer-reviewed & Survey on 283 AI incidents \\
  \citet{critch_tasra_2023} & Societal-scale AI risks & Public & Pre-print & Focus on accountability \& fault tree analysis \\
  \citet{weidinger_taxonomy_2022} & Large language models & AI community & Peer-reviewed & Observed vs anticipated risks \\
  \citet{bommasani_opportunities_2022} & Large language models & AI community & Pre-print & Survey on opportunities \& risks \\
  Huitri et al. (2024) & Generative speech models & Policy makers & Pre-print & Affected vs responsible entity \\
  Giudici et al. (2023) & AI risk monitoring in finance & Industry & Peer-reviewed & Statistical methods \\
  Golpayegani et al. (2022) & AI risk definition in health & Industry & Peer-reviewed & Machine-readable taxonomy \\
  Golpayegani et al. (2023) & AI risk taxonomy & AI community & Peer-reviewed & Open vocabulary (on EU AI Act) \\
  \end{tabular}
  \small
  \caption{Overview of Existing AI Risk and Harm Taxonomies}
  \label{tab:existing_taxonomies}
  \end{table*}

Risk and harm taxonomies are at the heart of risk management practices in industries such as aerospace, defense, and healthcare. They constitute the basis for risk modelling and mitigation methods, e.g.: Failure Mode and Effects Analysis (FMEA) \citep{rismani2023plane,koessler2023risk}. Centuries of engineering convey the lesson that having a common understanding and vocabulary of any technological harm is paramount for its identification and evaluation, and help inform inform policy-making and regulatory compliance.

Various works have attempted to characterise the risks and harms of AI and related systems.
Some are generic in scope, not limited to a particular domain or technology, while others focus on specific sectors or to  particular technology lifecycle stages.
\Cref{tab:existing_taxonomies} provides a summary of these taxonomies emphasising their purpose/focus, target audience, and features.

\subsection{Generic Taxonomies}
Mostly published by inter-governmental organisations of different kinds, generic taxonomies tend to focus on incident analysis and reporting, risk management, and ethical design.

\paragraph{Incident Analysis \& Reporting:}
An AI classification system~\cite{oecd_stocktaking_2023}  developed by The Organisation for Economic Co-operation and Development (OECD) categorises AI systems across five dimensions: people \& planet, economic context, data \& input, AI model, and task \& output.
Aimed primarily at policymakers, the framework aims at helps understand the implications of AI systems and their alignment with the OECD AI Principles.

The Center for Security and Emerging Technology (CSET) AI Harm framework \cite{cset} characterises harm largely from a safety perspective. It differentiates between tangible and intangible harms, and potential \emph{vs} realised harms. It also provides a customisation framework with guidelines on how to adapt it for a specific use. The Responsible AI Collective's AI Incident Database (AIID) appears to be an example of one such customisation \cite{mcgregor2021preventing}.\looseness=-1

\paragraph{Risk management:} The National Institute of Standards and Technology (NIST) drafted a Taxonomy of AI Risk,\footnote{\url{https://www.nist.gov/system/files/documents/2021/10/15/taxonomy_AI_risks.pdf}} which is part of the Artificial Intelligence Risk Management Framework~\cite{nist_risk}.
Inspired by three relevant policy documents (OECD\footnote{\url{https://www.oecd.org/going-digital/ai/principles}}, European Union Digital Strategy's Ethics Guidelines for Trustworthy AI,\footnote{\url{https://digital-strategy.ec.europa.eu/en/library/ethics-guidelines-trustworthy-ai}} and the US Executive Order 13960\footnote{\url{https://www.federalregister.gov/documents/2020/12/08/2020-27065/promoting-the-use-of-trustworthy}}), the NIST proposes a hierarchical taxonomy divided into three broad categories: Technical design attributes, Socio-technical attributes, and Guiding principles contributing to trustworthiness. The taxonomy aims to promote collaboration between the AI community, encouraging consensus on risk-related terminology.

\paragraph{Ethical design:}
Other harm taxonomies are intended to support the responsible and/or ethical design and development of AI systems, such as  one by the Alan Turing Institute \cite{alanturing}.
Similarly, Microsoft has published guidelines for general technology harm modeling for product teams \cite{microsoft}, which promotes adaptations for personal use.

\subsection{Specific Taxonomies}
The surveyed specific or focused taxonomies are mostly research articles and pre-prints, which can be differentiated in harm types, model types, and specific industries.

\paragraph{Specific harm type:}
Other taxonomies focus on specific types of harms, include one focusing on the socio-technical harms of algorithmic systems \cite{shelby_sociotechnical_2023} based on surveying 172 papers.
\citet{raji_fallacy_2022} analysed incidents from AIAAIC to propose an AI system failure taxonomy to connect them with actual harms.
TASRA is scoped around societal-scale risks and accountability \cite{critch_tasra_2023}. The authors based it on a fault tree analysis, resulting in a complex 6-level taxonomy.

\paragraph{Specific model types:} Other taxonomies focus on specific model types such as large language models \cite{weidinger_taxonomy_2022}, calling for extensions through case studies or interviews.  \cite{bommasani_opportunities_2022} also focuses on large language models (or foundational models as they coin them) and provide a very detailed survey on related work but do not attempt a comprehensive risk or harm taxonomy.
\citet{epic} provides an overview of the risks of generative AI with illustrative case study examples without claiming comprehensiveness.
Another taxonomy by researchers from Sony focuses on generative AI focused on speech \cite{hutiri2024not}, differentiating between affected and responsible entities.

\paragraph{Industry-specific:}
\citet{safe} provides an overview of the risks of AI in line with recent regulations and frameworks, such as the EU AI Act, and proposes a related risk measurement model, focused on the financial domain. \citet{safe2} extends the model to further domains, providing reproducible Python code.
\citet{golpayegani2022towards} also developed a taxonomy of AI risks in the health domain. In the context of EU AI Act, the vocabulary of AI risks (VAIR) \citet{golpayegani2023high} provides a formal taxonomy of concepts related to AI systems and their risks.
Finally, \citet{burema2023sector} analyses 125 incidents involving AI systems across five sectors (police, education/academia, politics, healthcare, automotive) to understand how ethical principles are breached in sector-specific contexts, finding that while some ethical issues span across sectors, others are sector-specific and relate to pre-existing structures and activities within those sectors.

We conclude that most existing taxonomies have different foci in terms of target audience and features, and accordingly mean different things to different people.

Further, only a few take an 'outside-in', holistic view of the actual, real-world impacts of on individuals, communities, businesses, society the environment, and other entities.

    \label{tab:existing_spec_taxonomies}

\section{Methodology}
\label{sec:methodology}

\subsection{Working Group Setup And Core Aims}

The proposed harms taxonomy was developed using an open, collaborative and structured process.
Its design was informed by more than 1,500 incidents and issues documented in the AIAAIC database. These real-world cases are based on an estimated 10,000+ media, research, legal and other reports from all around the globe, making the AIAAIC one of the largest public records of harms driven by AI and algorithmic systems, but one which was becoming difficult to manage due to limitations of its current taxonomy\footnote{\url{https://www.aiaaic.org/aiaaic-repository/classifications-and-definitions}} and the emergence of new risks and harms due to generative AI, emotion recognition and other technologies and applications.

This taxonomy is the result of the collective effort of an independent working group. Since June 2023, 25+ volunteers from various backgrounds, nationalities, genders and levels of seniority, replied to an open call for participation, all of whom taking part in a personal capacity to develop a general purpose, outside-in, open taxonomy of near-term risks and harms of AI, algorithms, and automation\footnote{List of individuals involved are listed in \hyperref[sec:acknowledgments]{Acknowledgments}}.

Unlike many of the taxonomies previously mentioned, the size, variety and openness of the working group ensured a maximal level of transparency, independence and diversity, both in the elaboration of the methodology and the taxonomy. The working group gathered at least once every two weeks, took decisions based on rough consensus\footnote{As defined by the IETF in RFC 7282 \url{https://datatracker.ietf.org/doc/html/rfc7282}.}, and settled on a set of core objectives for the taxonomy, notably:

\begin{itemize}
    \item Human-centred, based on actual incidents and their impact on different existing stakeholders
    \item Clear and understandable, both to domain experts, policy-makers and the general public
    \item Practical and accessible to civil society organisations, including news publishers, policy-makers, activists, businesses and the general public
    \item Comprehensive and applicable to harms stemming from the use of a wide variety of AI, algorithmic or algorithmic technologies and applications
    \item Inclusive of various differing social, geographical, and cultural contexts
    \item Flexible and extensible, to be easily updated to account for technological developments, adoption, and the emergence of new use cases
    \item Machine-readable to facilitate further development and encourage adoption, including for compliance purposes\footnote{For instance, as part as the "quality management systems" for high risk AI systems required by Article 17 of the EU AI Act }.
\end{itemize}
These core aims are also reflected in the working methods of the working group.

\subsection{Elaboration, Testing And Improvement Process}
The harms taxonomy was developed and refined using a structured, iterative methodology that combines a variety of approaches.

\subsubsection{Use case analysis:} The working group first mapped the needs and set the initial goals of the taxonomy through an audit of AIAAIC Repository users. Approximately 50 users, comprising working group members and others identified as having taken part in the AIAAIC's 2023 user survey were surveyed by email on their use of the Repository, what they saw as its limitations and opportunities, and their requirements. Participants were individual users participating in a personal capacity from organisations including Commonwealth Scientific and Industrial Research Organisation (CSIRO), Mozilla, University College London, and the University of Science and Technology of China.

\subsubsection{Literature review:} Relevant existing harms taxonomies were assessed to identify similarities, gaps, strengths and limitations. These previous works are mostly mentioned above and an overview of their respective foci, target audiences and features are outlined in Table 1.

This initial mapping, along with a review of current literature exploring the risks and harms of AI and related technologies~\cite{council2023ghost,hendrycks2023overview,bengio2023managing,tobin2023artificial}, informed the elaboration of a first version of the taxonomy, deliberately focused on near-term risks and documented harms of AI technologies. Key human rights and civil liberties charters, including the UN Universal Declaration of Human Rights, the European Convention on Human Rights, the Charter of Fundamental Rights of the European Union as well as the European Declaration on Digital Rights and Principles for the Digital Decade were also taken into account to produce the initial definition of the harms.

\subsubsection{Expert outreach:} This initial ``work-in-progress'' version of the taxonomy was presented to experts at NGOs, universities, news publishers and other organisations across the world to collect feedback on its thoroughness and practicality, especially on topics including Economics, Human rights and civil liberties, Misinformation and disinformation, Law, Politics, and Sustainability. The experts who provided feedback are acknowledged and thanked at the end of this paper.

\subsubsection{Annotation testing:} By selecting and annotating entries on the AIAAIC Repository, either randomly or choosing cases evidently different, and annotating them, the working group tested and refined the initial taxonomy to ensure it was clear, easy to use, and extensible.  Individual annotations were then compared to identify cases with diverging classifications, collect feedback on the definitions and adapt incrementally the taxonomy, week after week, to maximise consensus and efficiency.

\begin{figure}
    \centering
    \includegraphics[width=1\linewidth]{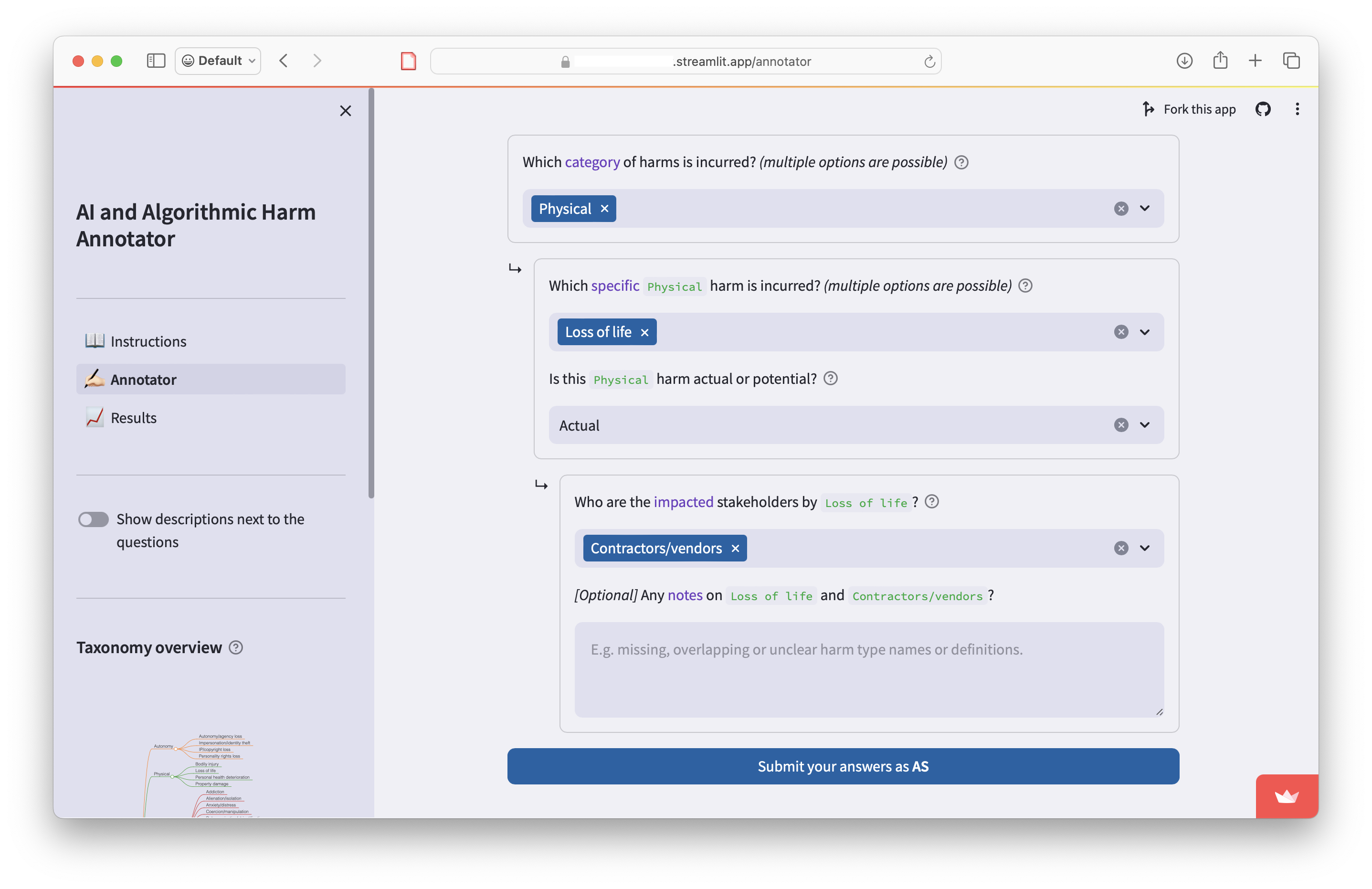}
    \caption{A screenshot of the annotation testing tool, tailored for collectively assessing and improving the taxonomy.}
    \Description{A screenshot of the annotation tool, showcasing the active 'Annotator' section. The left pane provides navigation links to 'Instructions,' 'Annotator,' and 'Results,' while the right pane displays a series of classification questions to be answered.}
    \label{fig:annotation_tool}
\end{figure}

To facilitate this process, a dedicated open-source tool was developed by the working group. This user-friendly interface enables quick annotation through drop-down menus with reminders of the definitions, streamlined comment submission, faster human review and consensus assessment. This tool also generates, for each annotated incident, a Sankey diagram, visually depicting the annotators' answers for each incident and computes its Krippendorff's alpha coefficient, a score measuring agreement--or lack thereof--among annotators \cite{gwet2014handbook}\footnote{The average alpha coefficient is expected to increase progressively towards 1, indicating perfect agreement, as the taxonomy improves. However, it will likely never reach 1 due to standard deviation from crowdsourced annotations and the emergence of new use cases. This provides the working group with a clear KPI.}, allowing for a swift and objective identification of potential gaps, classification conflicts or contentious definitions.
During the 9 first rounds of annotation, from February to April 2024, the working group generated over 1,000 individual annotations on 39 incidents pulled from the AIAAIC Repository.

\subsubsection{Validation through broader community review} This methodology of continuous incremental improvement and adaptation through multiple, parallel approaches is set to continue as an ongoing process. The methodology and tools set up by the working group are meant to scale and support a roadmap towards a more open and inclusive contribution process, involving a larger community of contributors, both for general or specialised reviews of the taxonomy itself, the annotation of historic cases in existing databases, or the ``live'' classification of new incidents upon reporting them or documenting them in harms and risks databases.

\section{Overview of the Harms Taxonomy} \label{sec:taxonomy}

\begin{figure*}[t]
    \centering
    \includegraphics[width=\linewidth]{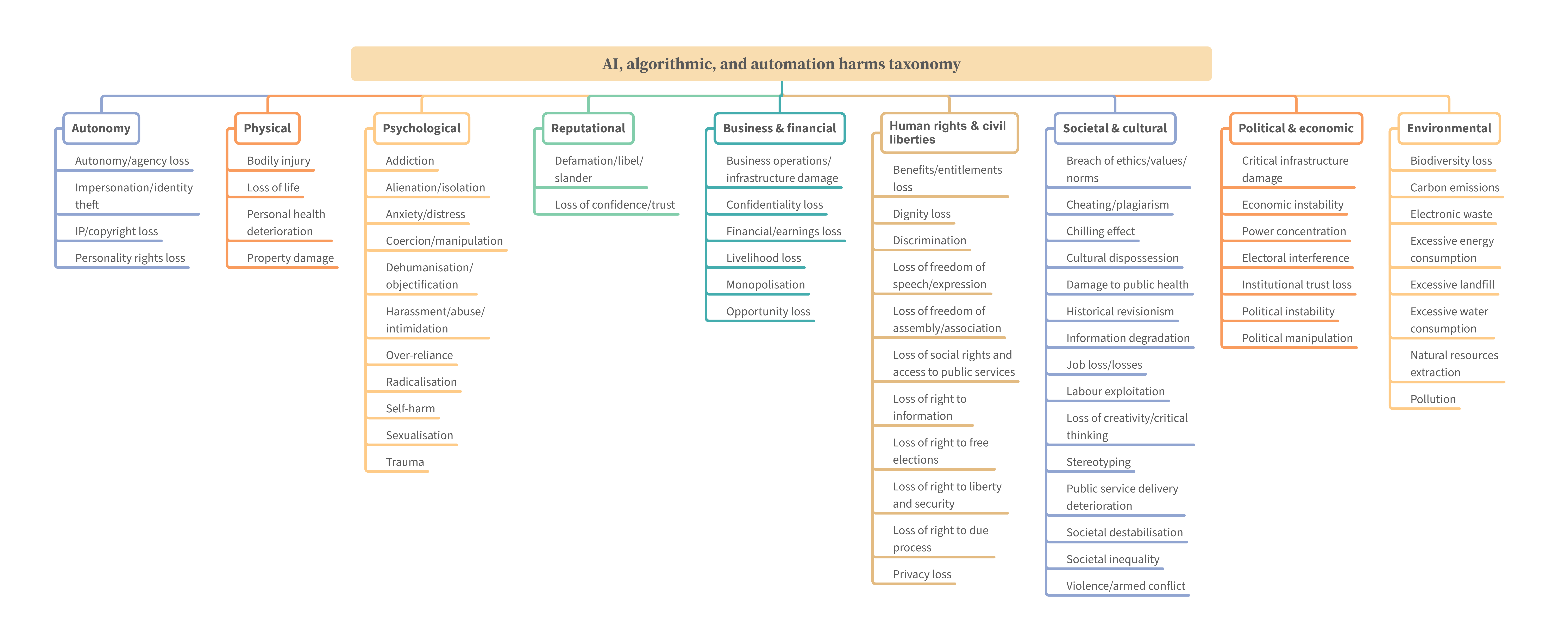}
    \caption{An overview of the AI, algorithmic and automation harms taxonomy.
    A printer-friendly version is available in \cref{fig:enter-label-big} in \cref{sec:appendix_a}}.
    \Description{A tree diagram illustrating the hierarchical structure of the AI, algorithmic, and automation harms taxonomy. It displays the main categories and their respective subcategories.}
    \label{fig:enter-label}
\end{figure*}

The result of the research, annotation and external reviews outlined in the previous section is presented here. Focusing on harms, as distinguished from causes or risks, supports the project's aim to develop a taxonomy that is understandable and usable by a wide range of stakeholders. It also ensures that the taxonomy is not dependent on technology or context of use and can be applied to a wide range of technologies and applications.\looseness=-1

It is important to note that the taxonomy presented here is not intended to be static but is dynamic and will evolve through continued annotation efforts, external reviews, and analysis. We therefore share this first version of the taxonomy to communicate our findings to a wider audience and start the process of soliciting additional feedback for further improvements.

For the purposes of this taxonomy, a harm is defined as a ``physical, psychological or other form of damage to third-parties'' that results from the use of an AI, algorithmic or automation system. The taxonomy organises harms at two different levels of granularity. The first level is the ``harm type'', which describes the general category of harm. The second level divides harm types into sub-categories, referred to as ``specific harms''. Definitions are provided for each harm category and specific harm.

The taxonomy outlines nine harm types:
\begin{itemize}
    \item \textbf{Autonomy} - Loss of or restrictions to the ability or rights of an individual, group or entity to make decisions and control their identity and/or output.
\end{itemize}
\begin{itemize}
    \item \textbf{Physical} - Physical injury to an individual or group, or damage to physical property.
\end{itemize}
\begin{itemize}
    \item \textbf{Psychological} - Direct or indirect impairment of the emotional and psychological mental health of an individual, organisation, or society.
\end{itemize}
\begin{itemize}
    \item \textbf{Reputational} - Damage to the reputation of an individual, group or organisation.
\end{itemize}
\begin{itemize}
    \item \textbf{Financial and Business} - Use or misuse of a technology system in a manner that damages the financial interests of an individual or group, or which causes strategic, operational, legal or financial harm to a business or other organisation.
\end{itemize}
\begin{itemize}
    \item \textbf{Human Rights and Civil Liberties} - Use or misuse of a technology system in a manner that compromises fundamental human rights and freedoms.
\end{itemize}
\begin{itemize}
    \item \textbf{Societal and Cultural} - Harms affecting the functioning of societies, communities and economies caused directly or indirectly by the use or misuse technology systems.
\end{itemize}
\begin{itemize}
    \item \textbf{Political and Economic} - Manipulation of political beliefs, damage to political institutions and the effective delivery of government services.
\end{itemize}
\begin{itemize}
    \item \textbf{Environmental} - Damage to the environment directly or indirectly caused by a technology system or set of systems.\looseness=-1
\end{itemize}
A diagram of specific harms, organised by harm type, is shown in \hyperref[fig:annotation_tool]{Figure 1}. The definitions for specific harms can be found in \hyperref[sec:appendix_a]{Appendix A}.

\begin{table*}
\footnotesize
\centering
  \begin{tabular}{
    p{3.1cm}
    >{\centering\arraybackslash}p{1cm}
    >{\centering\arraybackslash}p{1cm}
    >{\centering\arraybackslash}p{0.8cm}
    >{\centering\arraybackslash}p{0.8cm}
    >{\centering\arraybackslash}p{1.8cm}
    >{\centering\arraybackslash}p{1cm}
    >{\centering\arraybackslash}p{1.2cm}
    >{\centering\arraybackslash}p{0.8cm}
    >{\centering\arraybackslash}p{0.8cm}
  }
      & AIAAIC & OECD & EPIC & CSET & Alan Turing Institute & Microsoft Azure & Sony & TASRA & SHAS \\
    Autonomy
      & \checkmark & & \checkmark & & & & \checkmark & & \checkmark \\
    Physical
      & \checkmark & \checkmark & \checkmark & \checkmark & & \checkmark & \checkmark &  & \checkmark \\
    Psychological
      & \checkmark & \checkmark & \checkmark & & \checkmark & \checkmark & \checkmark & \checkmark & \checkmark \\
    Reputational
      & \checkmark & \checkmark & \checkmark & & & & \checkmark &  & \\
    Financial \& Business
      & \checkmark & & \checkmark & \checkmark & & & \checkmark & \checkmark & \checkmark\\
    Human Rights \& Civil Liberties
      & \checkmark & \checkmark & \checkmark & \checkmark & \checkmark & \checkmark & & & \checkmark \\
    Societal \& Cultural
      & \checkmark & \checkmark & & \checkmark & \checkmark & & \checkmark & \checkmark &\checkmark \\
    Political \& Economic
      & \checkmark & \checkmark & \checkmark & \checkmark & & \checkmark & \checkmark &\checkmark & \checkmark \\
    Environmental
      & \checkmark & & & \checkmark & & \checkmark & & &\checkmark \\
  \end{tabular}
    \caption{Comparison of harms taxonomies}
  \label{tab:taxonomycomparison}
\end{table*}

Though the harm types presented here are intended to be reasonably comprehensive, we recognise that new harm types will almost certainly have to be added in the future. Nonetheless, we expect that future additions to the taxonomy are more likely to occur at the specific harms level rather than the harm types level.

The specific harms are not intended to be mutually exclusive. Similar specific harms may exist within two or more harm types as necessary. For example, personal health deterioration is a specific harm within the Physical harm type, while damage to public health is a specific harm within the Societal and Cultural harm type. Where a similar harm exists in multiple harm types, they are intended to represent distinct concepts (e.g., personal health deterioration is focused on the harm to individual health while damage to public health is focused on harm to groups or communities). However, specific harms may be related; when personal health deterioration affects a large number of individuals, damage to public health is also likely to occur.

Additionally, we do not envision restricting the relationship between a particular incident and the harm to be one-to-one. A single incident may result in multiple harms, including multiple different specific harms within the same harm category. Within the custom tool used for annotation testing, annotators are allowed to list as many harms as they have identified for a particular incident.

Finally, the taxonomy envisions that annotators will characterise each identified harm as either ``actual'' or ``potential'' with respect to the particular incident being annotated. A harm is considered actual when the articles, reports and other public sources of information about the incident explicitly state that the harm has occurred. A harm is potential when the articles, reports and other sources of information about the incident indicate or suggest that a harm is reasonably likely to occur.

To facilitate a comparative understanding of our proposed taxonomy, we have selected other prominent taxonomies based on their scope, purpose, and intended audience, ensuring a relevant and meaningful comparison. The comparison, outlined in Table \ref{tab:taxonomycomparison}, features frameworks developed by recognised institutions such as the OECD, EPIC, CSET, the Alan Turing Institute, Microsoft Azure, Sony, and SHAS \cite{rismani2023plane}. These taxonomies were chosen because they similarly address a broad range of AI-related harms and are designed for diverse uses including policy guidance, risk management, and public education, and are intended for a wide audience encompassing policymakers, industry stakeholders, and the general public. Each column in the table represents a different taxonomy and indicates whether specific types of harm are addressed within their frameworks, highlighting how the AI harms taxonomy aligns with or differs from established approaches in the field. Among the taxonomies we compared, those developed by the OECD, CSET, and SHAS are most similar to ours due to their comprehensive scopes and focus on both policy-oriented applications and broad societal impacts. These frameworks emphasise a wide coverage of harms, both tangible and intangible, aligning closely with our objective to inform risk management and policy guidance comprehensively. The SHAS taxonomy, in particular, shares our emphasis on socio-technical aspects, making it highly relevant for interdisciplinary approaches to understanding and mitigating AI-related harms.

\section{Benefits and Intended Uses}
\label{uses}

The taxonomy proposed in this paper is intended to make the harms of AI, algorithmic and automation systems understandable and usable to a wide range of audiences, from experts to the general public.

We believe the taxonomy can be used for a wide variety of purposes.

\paragraph{Literacy and education:} Research studies regularly show a significant gap between awareness and understanding of AI systems---a gap that is reportedly widening, leading to growing concerns about the use of the technology in daily life in the US,\footnote{\url{https://www.pewresearch.org/short-reads/2023/08/28/growing-public-concern-about-the-role-of-artificial-intelligence-in-daily-life/}} UK,\footnote{\url{https://www.gov.uk/government/publications/public-attitudes-to-data-and-ai-tracker-survey-wave-3/public-attitudes-to-data-and-ai-tracker-survey-wave-3}} and elsewhere.

Despite these concerns, and the opaque and high-risk nature of some algorithmic systems, governments, educators and others recognise that these technologies play an increasingly central role in the everyday lives of citizens and consumers, government, civil society and business. As such, it is important that they are  demystified and made understandable to everyone.

This realisation is evident in the use of the AIAAIC Repository by teachers, academics, professional associations, businesses and other entities to educate decision-makers, students, employees, suppliers and others on the nature, risks and harms of AI, algorithmic and automation systems \cite{bussell2023scaffolding}.

The harms taxonomy should help make the full range of AI and algorithmic harms of clearer and easier to understand by a broader set of audiences.

\paragraph{Journalism:} AI and algorithmic decision-making for public or commercial purposes are of significant interest to journalists and other civil opinion-formers. However, it can be difficult to untangle actual impacts from official and unofficial sources, each of which may be partial and skewed.

By providing evidence-based, impartial data on the use and misuse of AI, algorithmic and automation systems, incident databases incorporating the ability to sort a wide range of actual and potential harms in an easy to understand and use manner can be of real value to technology, business, health and other journalistic beats, as well as to investigative journalists.\looseness=-1

Structured harms data can be used to inform journalistic enquiries and make the case to editors, in addition to supporting and strengthening news, commentary and analysis, and to create visual storytelling or ‘data journalism’.

\paragraph{Advocacy:} Non-governmental civil society organisations (NGOs) making the case for human rights and civil liberties, consumers, patients and other constituencies often expose abuses and violations in order to put pressure on perceived culprits--typically system developers and/or operators, but also malicious bad actors--and to make the case for stronger or better enforced legislation.

Sometimes, an NGO’s findings and/or third-party accusations are recorded as a public resource---an example being the Business \& Human Rights Resource Centre’s news database.\footnote{\url{https://www.business-humanrights.org/en/latest-news/}}

However, the majority of civil society organisations have few resources and struggle to record their findings in a structured manner, or maintain databases. The harms taxonomy can help NGOs identify common harms across different systems, countries and cultures, strengthen existing databases, and create new systems to track harms and related information.\looseness=-1

\paragraph{Citizen reporting:} The ability for citizens and other users to hold the developers and deployers of AI and algorithmic technology systems are currently hampered by complaint reporting systems that are often challenging to find, cumbersome to use, and which can deliver slow and partial responses.

Furthermore, it is often not in the interests of these entities to disclose incidents reported by citizens and other users publicly, lest they be exposed to greater legal, financial, operational, and reputational risk.

The AIAAIC Repository and the AI Incident Database (AIID) \footnote{\url{https://incidentdatabase.ai/}} are considered to be the only two tools that enable individuals and organisations to report incidents based on media reports and, in the case of AIAAIC, legal dockets, research reports and other materials. However, the two entities use different taxonomies and definitions for defining harms, resulting in inconsistent outputs\footnote{\url{https://cset.georgetown.edu/wp-content/uploads/CSET-An-Argument-for-Hybrid-AI-Incident-Reporting.pdf}}.

Impending legislation, notably in Brazil, Canada and the European Union\footnote{\url{https://www.europarl.europa.eu/news/en/press-room/20240308IPR19015/artificial-intelligence-act-meps-adopt-landmark-law}}, provides citizens with the right to submit complaints about high-risk AI systems and to receive explanations about how these systems work.

Governments adopting these proposals will need to develop reporting systems that clearly spell out the harms associated with these systems so that citizens, consumers, patients, etc, can report incidents and issues in a clear, user-friendly and consistent manner.

The harms taxonomy can help governments and government agencies develop user-friendly, effective reporting systems that align user needs and expectations with their own obligations and requirements.

Furthermore, the taxonomy may prove useful in developing a federated and standardised hybrid reporting framework of the kind proposed by the Center for Security and Emerging Technology\footnote{\url{https://cset.georgetown.edu/wp-content/uploads/CSET-An-Argument-for-Hybrid-AI-Incident-Reporting.pdf}}.

In addition, civil society organisations such as advocacy groups, news publishers and other third-party watchdogs can use the taxonomy to inform reports and investigations, and to develop databases and systems tracking human rights, political, environmental and other misuses and violations.

\paragraph{Policy making and enforcement:} Databases such as the AIAAIC Repository and the OECD AI Incident Monitor\footnote{\url{https://oecd.ai/en/incidents}} provide policymakers and regulators with insights into the risks and harms of technological systems, enabling them to advise politicians and governments using data and examples, and to help set priorities for regulation and enforcement.

It may also help regulators develop taxonomies and systems that enable them to identify and track violations relevant to their remit.

\paragraph{Risk management:} The taxonomy proposed in this paper can also be used for risk management, auditing and impact assessment purposes. Several regulations and standards on AI systems underline the need to develop effective risk management models for AI systems (e.g. EU AI Act~\cite{aiact}, US NIST risk management framework \cite{nist_risk}).

Such models should be based on both the likelihood and the expected harm of an incident. While the former can be measured on the basis of statistical metrics that measure the compliance of AI to a set of Safe and trustworthy requirements (see e.g. \cite{safe2} for a consistent set of such metrics), the latter require an assessment of all possible harms that non-compliance can lead to. Traditional taxonomies focus on the harms caused to a private or public organisation, or to a national entity; however, our taxonomy is able to capture a wide spectrum of harms, occurring to individuals, organisations, society and the environment, across the  world, and is therefore better equipped to estimate risks, and prioritise consequent mitigation actions.\looseness=-1

The proposed taxonomy acts as an open and accessible guiding resource for compliance with risk management and fundamental rights impact assessment (FRIA) obligations of the EU AI Act. Given the cost burden of legal compliance, this is particularly useful for small and medium-sized businesses and less well resourced organisations providing or deploying AI and algorithmic systems.

Given the taxonomy will be regularly updated to reflect emerging harms, it should also serve as a valuable reference in shaping governance frameworks by providing insights on the current landscape of risks for policy-makers, regulators, and standardisation bodies.

\section{Conclusion and Future Work}
\label{sec:conclusion}

\textbf{Limitations:} The taxonomy is designed to be an ever-evolving work, and is still in a relatively early stage of development. Accordingly, it currently has a number of limitations that will be worked on and developed.

The creation of the taxonomy has been a labour-intensive endeavour, involving dozens of people with expertise over a wide variety of fields and subject areas, and has, to this point consumed hundreds of person-hours of work. This may not be sustainable in the long term, and yet it is important to ensure that relevance and coverage is maintained and that the taxonomy remains usable and useful to a broad range of users.

In addition, whilst the taxonomy has undergone multiple rounds of internal testing (see Methodology), it has not yet been tested with people outside the internal working group, and it remains to be seen how usable it is for people less familiar with its development.

One of the objectives of the development methodology for the taxonomy has been \textit{clarity}. The elements of the taxonomy including definitions has been provided to the annotators.  A visual hierarchical diagram of the taxonomy is always visible to annotators completing the task of annotation increasing clarity of the overall structure. Post-annotation, the resulting differences  have been explained through visual rendering of choices made by each individual annotator.  This visual element of pre-annotation structure and post-annotation choice graphs has resulted in greater clarity of the taxonomy and its use, but this can always be improved and made more relevant to different cultures and countries.

Another objective of the taxonomy is \textit{flexibility}. As the AI and algorithmic applications included in the database increase in scope and variety, the required classification system for harms i.e. the taxonomy must necessarily evolve.  In some cases, there will be new categories of harm that are not covered adequately by the current taxonomy. In other cases the definitions and grouping of categories will need to be modified to capture emerging and growing areas of harm. The taxonomy therefore needs to be flexible enough to cover both the breadth and depth of an evolving set of harms from new technology. Whereas stability of a taxonomy is a desirable feature for some taxonomies, e,g, the base taxonomies for naming flora and fauna, in the area of classifying harm related to new and emerging technology flexibility is a required feature.

\textbf{Benefits:} The development of this taxonomy is \textit{human centred} and volunteer-based. The use of taxonomies to classify content, and in particular annotation has a dark side.  There are examples of large technology companies exploiting gig workers for the completion of annotation tasks, including the classification of harmful content \cite{A733755113}. The taxonomy developed here, and to be used for classifying and annotating events, is human centred in the sense not just that it is intended to be understood and used by end users as well as experts, but also in the sense that it is based on volunteer input as opposed to an exploitative employment contract.\looseness=-1

Driven by news, legal, research and other negative reports of AI and algorithmic events, the taxonomy has an \textit{outside-in} focus. By contrast, an inside-out taxonomy would develop a safety/vulnerability-focused taxonomy and then apply it from the first reported event onwards.  The shape of the outside-in taxonomy described here enabled the collection of a critical mass of externally reported events first, and then developed an internal taxonomy based on that database of events. The taxonomy will be modified based on significant changes to the underlying database of events.

The taxonomy is developed by a \textit{multidisciplinary} group of volunteers, including journalists, law, sociology, user experience, computer science and other experts from a number of different countries and levels of expertise and seniority.

The development of this taxonomy has been \textit{independent} of financial support form  companies or governments. As described by \citet{AbdallaMohamed2021TGHP}, projects funded by big tech can be significantly oriented towards positive evaluations of the technology projects being researched, and minimisation of risks of the technology. \citet{GoldenfeinJake2023Tmic} indicates that a similar conflict of interest issue potentially exists among digital rights civil society organisations, which are disproportionately funded by big tech.  The taxonomy development process presented here has not received any finding from big technology companies, and is therefore relatively independent of these potential conflict of interest influences.\looseness=-1

The proposed taxonomy has also benefited from its data-driven design, which incorporates information from a database comprising 1,500+ entries driven by and relating to AI, algorithmic and automation incidents and issues across the world. The diversity in the dataset substantially enhances the comprehensiveness and exhaustiveness of the categories and sub-categories delineated.

\textbf{Future work:} The proposed taxonomy will be strengthened to uncover potential deficiencies or oversights, enhance openness, transparency and accessibility, and account for emerging technologies and harms. Proposed milestones for the working group include:

The current taxonomy is mostly descriptive and focused on harms; it could be expanded to categorise associated causes and risks. Such an extension would have the potential to support advocacy, policy-making, risk management and governance but it would imply a more analytical and interpretative dimension. This requires further research.

The taxonomy and annotation tool are intended to act as an open citizen science initiative. This will involve various stakeholders in assigning specific harms to incidents and issues documented in the AIAAIC Repository and/or similar databases. Through such an initiative, a broader set of stakeholders will be empowered to democratise important knowledge relating to AI, algorithmic and automation harms.

\section{Ethical Statement}
\label{sec:ethical_statement}

The taxonomy presented here is designed to be an ever-evolving work in progress, and is still in a relatively early stage of development.  At the same time, there are ethical considerations for the ongoing use of the taxonomy.

\paragraph{Oversimplification:} The exercise of classifying harms is in itself an exercise in simplification. Although this exercise of classification has many benefits as outlined earlier, there is a risk of \textit{oversimplification} in the form of a lack of breadth in the taxonomy. For example the number of categories and structure of the taxonomy can oversimplify a complex phenomenon of technology-based, real-world harms. Using an oversimplified taxonomy can result in misunderstandings about the nature and severity of harms involved, resulting in inadequate responses or interventions.

\paragraph{Normalisation:} Categorising an event as having a particular type of harm can minimise or reduce its importance. Harms are often specific to a group or an individual, and can have highly negative consequences for those particular groups or individuals that is not experienced by broader groups. \textit{Normalisation} is a process of categorising an event in accordance with comparisons of harm from other events. The depth of the taxonomy determines the relative classification and categorisation of harms. Using a taxonomy that is not appropriately deep, is inappropriately scaled, and can result in mis-classification, resulting in inadequate scale of responses or interventions.

\paragraph{Ineffective Risk Mitigation:} Taxonomies are political projects and taxonomies can be used in ways that do not effect actual reduction in the risks and harms identified through classification. There is a risk of \textit{ineffectiveness} if the time and effort invested in developing and applying a taxonomy does not result in actual reduction in risks and harms. For example the number of categories and structure of the taxonomy can be mismatched to the way in which the technology systems are designed, and therefore taxonomy can be of limited use to developers and designers. Similarly, the taxonomy can be mismatched to the process by which civil society organisations advocate for the reduction of harms for the groups they represent. Using an ineffective taxonomy can result in a classification of events, but no consequent response, intervention or preventive process implementation.

\begin{acks}
  \label{sec:acknowledgments}
We are grateful to the following individuals for their input to discussions relating to AIAAIC’s harms taxonomy and to its broader Risks/harms taxonomy project:

\begin{itemize}
    \item Adriana Eufrosina Bora
    \item Alex Read
    \item Alice Poorta
    \item Amari Cowan
    \item Athena Vakali
    \item Bastian Cibbe
    \item Danny Rayman Labrin
    \item Pavlos Sermpezis
    \item Emma Ruttkamp-Bloem
    \item Henry Ajder
    \item Jared Katzman
    \item John Havens
    \item John-Patrick Akinyemi
    \item Laureen van Breen
    \item Matin Abdullah
    \item Mike Walton
    \item Parvati Neelakantan
    \item Samanti Wijeyesekera
    \item Sara Pido
    \item Tara Garcia Mathewson
    \item Tim Carter
    \item Vasiliki Gkatziaki
\end{itemize}

\end{acks}

\bibliographystyle{ACM-Reference-Format}
\bibliography{bibliography}

\appendix
\appendix

\section{Appendix A}
\label{sec:appendix_a}

This appendix provides definitions for each specific harm. The specific harms are organised by harm type. The definitions for harm types may be found in the Taxonomy section in the main body of the paper above.

\begin{landscape}
\begin{figure*}[t]
    \centering
    \includegraphics[width=\linewidth]{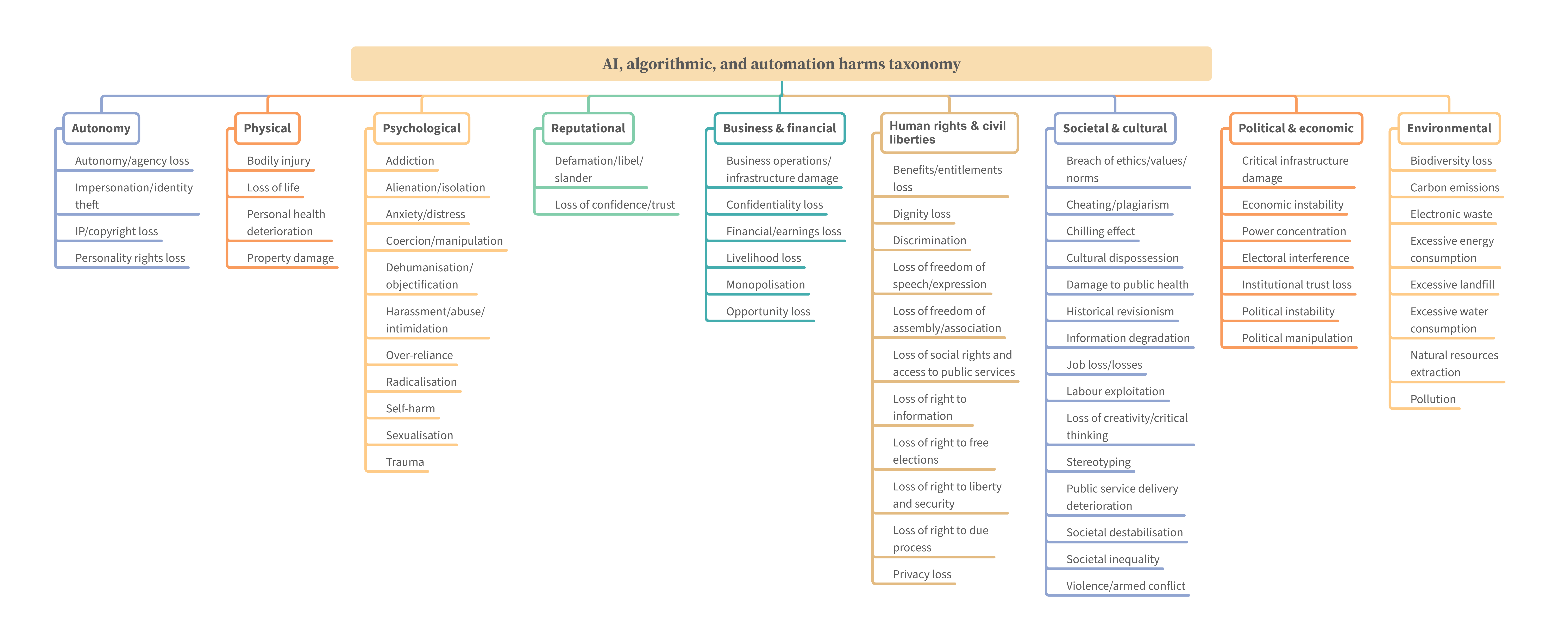}
    \caption{An overview of the AI, algorithmic and automation harms taxonomy}
    \Description{A tree diagram illustrating the hierarchical structure of the AI, algorithmic, and automation harms taxonomy. It displays the main categories and their respective subcategories.}
    \label{fig:enter-label-big}
\end{figure*}
\end{landscape}

For the printer-friendly version of the taxonomy diagram, please refer to \cref{fig:enter-label-big} at the end of the document.

\subsection{Autonomy}
\begin{description}
    \item[Autonomy/agency loss] - Loss of an individual, group or organisation’s ability to make informed decisions or pursue goals.
    \item[Impersonation/identity theft] - Theft of an individual, group or organisation’s identity by a third-party in order to defraud, mock or otherwise harm them.
    \item[IP/copyright loss] - Misuse or abuse of an individual or organisation’s intellectual property, including copyright, trademarks, and patents.
    \item[Personality rights loss] - Loss of or restrictions to the rights of an individual to control the commercial use of their identity, such as name, image, likeness, or other unequivocal identifiers.
\end{description}

\subsection{Physical}
\begin{description}
    \item[Bodily injury] - Physical pain, injury, illness, or disease suffered by an individual or group due to the malfunction, use or misuse of a technology system.

    \item[Loss of life] - Accidental or deliberate loss of life, including suicide, extinction or cessation, due to the use or misuse of a technology system.

    \item[Personal health deterioration] - Physical deterioration of an individual or animal over time, increasing their risk of disease, organ failure, prolonged hospital stay or death, etc.

    \item[Property damage] - Action(s) that lead directly or indirectly to the damage or destruction of tangible property eg. buildings, possessions, vehicles, robots.
\end{description}

\subsection{Psychological}
\begin{description}
    \item[Addiction] - Emotional or material dependence on technology or a technology system.

    \item[Alienation/isolation] - An individual's or group's feeling of lack of connection with those around as a result of technology use or misuse.

    \item[Anxiety/depression] - Mental health decline due to addiction, negative social interactions such as humiliation and shaming and traumatic distressing events such as online violence or rape.

    \item[Coercion/manipulation] - Use of a technology system to covertly alter user beliefs and behaviour using nudging, dark patterns and/or other opaque techniques, resulting in potential erosion of privacy, addiction, anxiety/distress, etc.

    \item[Dehumanisation/objectification] - Use or misuse of a technology system to depict and/or treat people as not human, less than human, or as objects.

    \item[Harassment/abuse/intimidation] - Online behaviour, including sexual harassment, that makes an individual or group feel alarmed or threatened.

    \item[Over-reliance] - Unfettered and/or obsessive belief in the accuracy or other quality of a technology system, resulting in addiction, anxiety, introversion, sentience, complacency, lack of critical thinking and other actual or potential negative impacts.

    \item[Radicalisation] - Adoption of extreme political, social, or religious ideals and aspirations due to the nature or misuse of an algorithmic system, potentially resulting in abuse, violence, or terrorism.

    \item[Self-harm] - Intentional seeking out or sharing of hurtful content about oneself that leads to, supports, or exacerbates low self-esteem and self-harm.

    \item[Sexualisation] - Sexual interest in a technology or application.

    \item[Trauma] - Severe and lasting emotional shock and pain caused by an extremely upsetting experience.
\end{description}

\subsection{Reputational}
\begin{description}
    \item[Defamation/libel/slander] - Use of a technology system to create, facilitate or amplify false perception(s) about an individual, group, or organisation.

    \item[Loss of confidence/trust] - Misleading or unfair change(s) in how an individual, group, or organisation is viewed, leading to loss of ability to conduct relationships, raise capital, recruit people, etc.
\end{description}

\subsection{Financial and Business}
\begin{description}
    \item[Business operations/infrastructure damage] - Damage, disruption, or destruction of a business system and/or its components due to malfunction, cyberattacks, etc.

    \item[Confidentiality loss] - Unauthorised sharing of sensitive, confidential information and documents such as corporate strategy and financial plans with third-parties.

    \item[Financial/earnings loss] - Loss of money, income or value due to the use or misuse of a technology system.

    \item[Livelihood loss] - An individual or group’s loss of ability to support themselves financially or vocationally due to natural disasters, lack of demand for products/services, cost increases, etc, resulting in inability to procure food, reduced employment prospects, bankruptcy, foreclosure, homelessness, etc.

    \item[Increased competition] - The inappropriate or unethical use of technology to gain market share.

    \item[Monopolisation] - Abuse of market power through the control of prices, thereby limiting competition and creating unfair barriers to entry.

    \item[Opportunity loss] - Loss of ability to take advantage of a financial or other opportunity, such as education, employability/securing a job.
\end{description}

\subsection{Human Rights and Civil Liberties}
\begin{description}
    \item[Benefits/entitlements loss] - Denial or or loss of access to welfare benefits, pensions, housing, etc due to the malfunction, use or abuse of a technology system.

    \item[Dignity loss] - Perceived loss of value experienced by or disrespect shown to an individual or group, resulting in self-sheltering, loss of connections and relationships, and public stigmatisation.

    \item[Discrimination] - Unfair or inadequate treatment or arbitrary distinction based on a person's race, ethnicity, age, gender, sexual preference, religion, national origin, marital status, disability, language, or other protected groups.

    \item[Loss of freedom of speech/expression] - Restrictions to or loss of people’s right to articulate their opinions and ideas without fear of retaliation, censorship, or legal sanction.

    \item[Loss of freedom of assembly/association] - Restrictions to or loss of people’s right to come together and collectively express, promote, pursue, and defend their collective or shared ideas, and/or to join an association.

    \item[Loss of social rights and access to public services] - Restrictions to or loss of rights to work, social security, and adequate standard of living, housing, health and education.

    \item[Loss of right to information] - Restrictions to or loss of people’s right to seek, receive and impart information held by public bodies.

    \item[Loss of right to free elections] - Restrictions to or loss of people’s right to participate in free elections at reasonable intervals by secret ballot.

    \item[Loss of right to liberty and security] - Restrictions to or loss of liberty as a result of illegal or arbitrary arrest or false imprisonment.

    \item[Loss of right to due process] - Restrictions to or loss of right to be treated fairly, efficiently and effectively by the administration of justice.

    \item[Privacy loss] - Unwarranted exposure of an individual's private life or personal data through cyberattacks, doxxing, etc.
\end{description}

\subsection{Societal and Cultural}
\begin{description}
    \item[Breach of ethics/values/norms] - An actual or perceived violation or deviation from the established societal values, norms or ethical standards or principles.

    \item[Cheating/plagiarism] - Use of another person’s or group’s words or ideas without consent and/or acknowledgement.

    \item[Chilling effect] - The creation of a climate of self-censorship that deters democratic actors such as journalists, advocates and judges from speaking out.

    \item[Cultural dispossession] - Intentional and/or unintentional erasure of cultural goods and values, such as ways of speaking, expressing humour, or sounds and voices that contribute to a cultural identity, or their inappropriate re-use in other cultures.

    \item[Damage to public health] - Adverse impacts on the health of groups, communities or societies, including malnutrition, disease and infection conditions.

    \item[Historical revisionism] - Deliberate or unintentional reinterpretation of established/orthodox historical events or accounts held by societies, communities, academics.

    \item[Information degradation] - Creation or spread of false, hallucinatory, low-quality, misleading, or inaccurate information that degrades the information ecosystem and causes people to develop false or inaccurate perceptions, decisions and beliefs; or to lose trust in accurate information.

    \item[Job loss/losses] - Replacement/displacement of human jobs by a technology system, leading to increased unemployment, inequality, reduced consumer spending, and social friction.

    \item[Labour exploitation] - Use of under-paid and/or offshore labour to develop, manage or optimise a technology system.

    \item[Loss of creativity/critical thinking] - Devaluation and/or deterioration of human creativity, artistic expression, imagination, critical thinking or problem-solving skills.

    \item[Stereotyping] - Derogatory or otherwise harmful stereotyping or homogenisation of individuals, groups, societies or cultures due to the mis-representation, over-representation, under-representation, or non-representation of specific identities, groups, or perspectives.

    \item[Public service delivery deterioration] - Poor performance of a public technology system due to malfunction, over-use, under-staffing etc, resulting in individuals, groups, or organisations unable to use it in a manner they can reasonably expect.

    \item[Societal destabilisation] - Societal instability in the form of strikes, demonstrations and other types of civil unrest caused by loss of jobs to technology, unfair algorithmic outcomes, disinformation, etc.

    \item[Societal inequality] - Increased difference in social status or wealth between individuals or groups caused or amplified by a technology system, leading to the loss of social and community wellbeing/cohesion and destabilisation.

    \item[Violence/armed conflict] - Use or misuse of a technology system to incite, facilitate or conduct cyberattacks, security breaches, lethal, biological and chemical weapons development, resulting in violence and armed conflict.
\end{description}

\subsection{Political and Economic}
\begin{description}
    \item[Critical infrastructure damage] - Damage, disruption to or destruction of systems essential to the functioning and safety of a nation or state, including energy, transport, health, finance, and communication systems.

    \item[Economic instability] - Uncontrolled fluctuations impacting the financial system, or parts thereof, due to the use or misuse of a technology system, or set of systems.

    \item[Power concentration] - Amplification of concentration of economic and/or political wealth and power, potentially resulting in increased inequality and instability.

    \item[Electoral interference] - Generation of false or misleading information that can interrupt or mislead voters and/or undermine trust in electoral processes.

    \item[Institutional trust loss] - Erosion of trust in public institutions and weakened checks and balances due to mis/disinformation, influence operations, over-dependence on technology, etc.

    \item[Political instability] - Political polarisation or unrest caused by increased inequality, job losses, over-dependence on technology making societies vulnerable to systemic failures, etc, arising from or amplified by the use or misuse of a technology system.

    \item[Political manipulation] - Use or misuse of personal data to target individuals’ interests, personalities and vulnerabilities with tailored political messages via micro-advertising or deepfakes/synthetic media.
\end{description}

\subsection{Environmental}
\begin{description}
    \item[Biodiversity loss] - Over-expansion of technology infrastructure, or inadequate alignment of technology with sustainable practices, leading to deforestation, habitat destruction, and fragmentation and loss of biodiversity.

    \item[Carbon emissions] - Release of carbon dioxide, nitric oxide and other gases, increasing carbon emissions, exacerbating climate change, and negatively impacting local communities.

    \item[Electronic waste] - Electrical or electronic equipment that is waste, including all components, sub-assemblies and consumables that are part of the equipment at the time the equipment becomes waste

    \item[Excessive energy consumption] - Excessive energy use, leading to energy bottlenecks and shortages for communities, organisations, and businesses.

    \item[Excessive landfill] - Excessive disposal of electrical or electronic equipment leading to ecological/biodiversity damage, and disrupting the livelihoods and eroding the rights of local communities.

    \item[Excessive water consumption] - Excessive use of water to cool data centres and for other purposes, leading to water restrictions or shortages for local communities or businesses.

    \item[Natural resource depletion] - Extraction of minerals, metals, rare earths, and fossil fuels that deplete natural resources and increase carbon emissions.

    \item[Pollution] - Actual or potential pollution to the air, ground, noise, or water caused by a technology system.
\end{description}

\end{document}